\documentclass{article}

\usepackage{arx_style}

\usepackage[utf8]{inputenc} 
\usepackage[T1]{fontenc}    
\usepackage{url}            
\usepackage{booktabs}       
\usepackage{amsfonts}       
\usepackage{nicefrac}       
\usepackage{microtype}      
\usepackage{xcolor}         
\usepackage{graphicx}
\usepackage{tikz}
\usepackage{amsmath}
\usepackage{adjustbox}
\newcommand\norm[1]{\lVert#1\rVert}
\usepackage{color, colortbl}
\usepackage{hyperref}       
\usepackage[capitalize,noabbrev]{cleveref}
\usepackage{enumitem}

\definecolor{Gray}{gray}{0.9}

\title{Group Orthogonalization Regularization for Vision Models Adaptation and Robustness}

\author{
    Yoav Kurtz \quad Noga Bar \quad Raja Giryes \\
    Department of Electrical Engineering \\ 
    Tel Aviv University
}

\begin{document}
\maketitle

\begin{abstract}  
As neural networks become deeper, the redundancy within their parameters increases. This phenomenon has led to several methods that attempt to reduce the correlation between convolutional filters. We propose a computationally efficient regularization technique that encourages orthonormality between groups of filters within the same layer. Our experiments show that when incorporated into recent adaptation methods for diffusion models and vision transformers (ViTs), this regularization improves performance on downstream tasks. We further show improved robustness when group orthogonality is enforced during adversarial training. Our code is available at \url{https://github.com/YoavKurtz/GOR}.
\end{abstract}

\begin{figure}[h]
\centering
\includegraphics[scale=0.62]{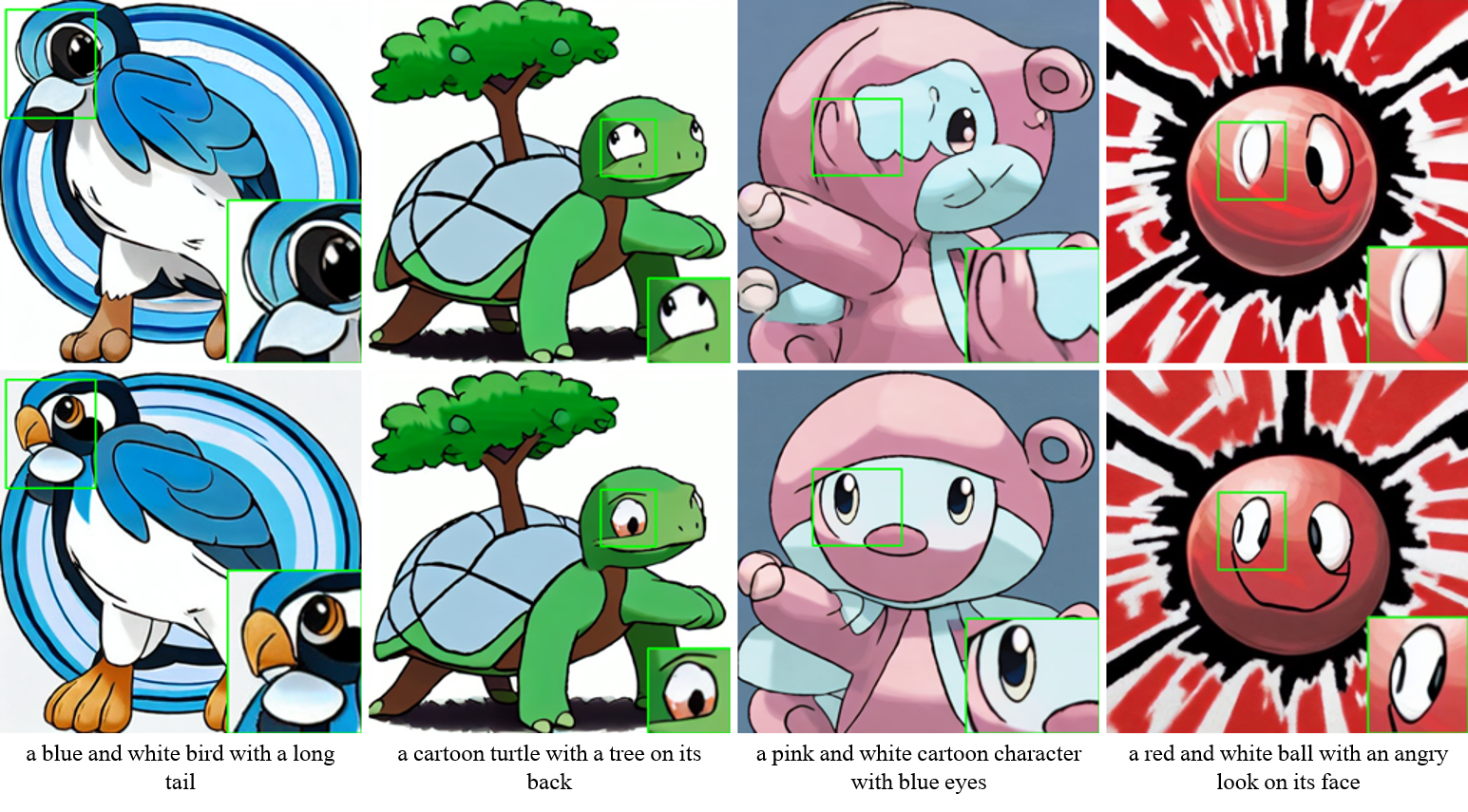}
\caption{Qualitative comparisons on Pokemon-BLIP between a baseline fine-tuned model using LoRA (top) and a model fine-tuned along with GOR (bottom) using the same seed. The green rectangle is zoomed in by a factor of $1.5$.  
Note the improved quality of GOR}
\label{fig:pokemon_crop_grid}
\end{figure}

\section{Introduction}

Deep neural networks have achieved remarkable success in various tasks, such as image classification and generation, object detection, and segmentation. However, as networks become deeper and wider, the redundancy within their parameters increases.
While increasing the DNN size can lead to improved expressivity and performance, it was shown that there is a high correlation between parameters \cite{wang2020orthogonal}.
This phenomenon has led to several methods that attempt to reduce the correlation between convolutional filters \cite{wang2020orthogonal,bansal2018can}.

Additionally, various methods are suggested to regularize the space of filters and features. One of the dominant examples is batch normalization \cite{bn}, which normalizes the activations of the previous layer to have zero mean and unit variance. For reducing the dependency of batch normalization on the batch size, it was suggested to use group normalization \cite{gn}, which normalizes for each example the activations of a subset of channels in a given layer to have zero mean and unit variance.

In addition to feature normalization, some methods have been proposed for normalizing the network weights. One example is weight standardization \cite{qiao2019micro}, which normalizes the weight vectors to have unit $\ell_2$ norm.
Despite their effectiveness, these methods either normalize each weight independently \cite{salimans2016weight,qiao2019micro} or all together \cite{xie2017all, bansal2018can,ozay2016optimization,jia2017improving}, which may be computationally demanding. To address this limitation, in our work, we address the correlation between filters in the same layer with reduced computational overhead. We propose an efficient regularization method that encourages orthonormality between filter groups within the same layer. Our approach, called Group Orthogonalization Regularization (GOR), is motivated by the observation that orthonormal filters are more diverse, expressive, and less redundant than correlated filters. Yet, instead of enforcing orthogonality between all filters, inspired by group normalization, we enforce orthogonality only between groups of filters. 

To assess the validity of our proposed method, we check it first on the well-known classification task of CIFAR-10 and show improvement in accuracy.

Moreover, as our regularization increases the expressivity of the model and imposes diversity on the filter, we use its advantages in extreme regimes where utilizing the expressive power of the model can be crucial for its performance. 
We evaluate GOR on recent low-rank adaptation methods for diffusion models \cite{ramesh2021zero,Saharia2022Photorealistic} and vision transformers (ViTs) \cite{dosovitskiy2020image}.
Specifically, we propose combining GOR with AdaptMLP \cite{chen2022adaptformer}, which is an adaptation strategy for ViT. We show that incorporating GOR into the AdaptMLP mechanism improves ViT classification accuracy on downstream datasets.
We also show how to combine GOR with LoRA \cite{hu2021lora} to adapt diffusion models. We demonstrate that this can lead to improvement in a variety of datasets (e.g., in Pokemon generation as shown in \cref{fig:pokemon_crop_grid}). This further illustrates our regularization's effectiveness.

Finally, we show that GOR is useful for improving network robustness. When group orthogonality is enforced during adversarial training, robustness is enhanced. Notably, our regularization method enables increasing layer dimensionality without sacrificing efficiency.

Overall, the main contribution of GOR is addressing the correlation between filters in the same layer by promoting orthonormality between filter groups.
Using our approach, layer dimensionality may be increased without compromising efficiency. 
We empirically demonstrate that our new regularization method complements existing normalization techniques and can be applied to a wide range of deep learning models and tasks.

\begin{figure}[t]
\centering
\includegraphics[scale=0.5]{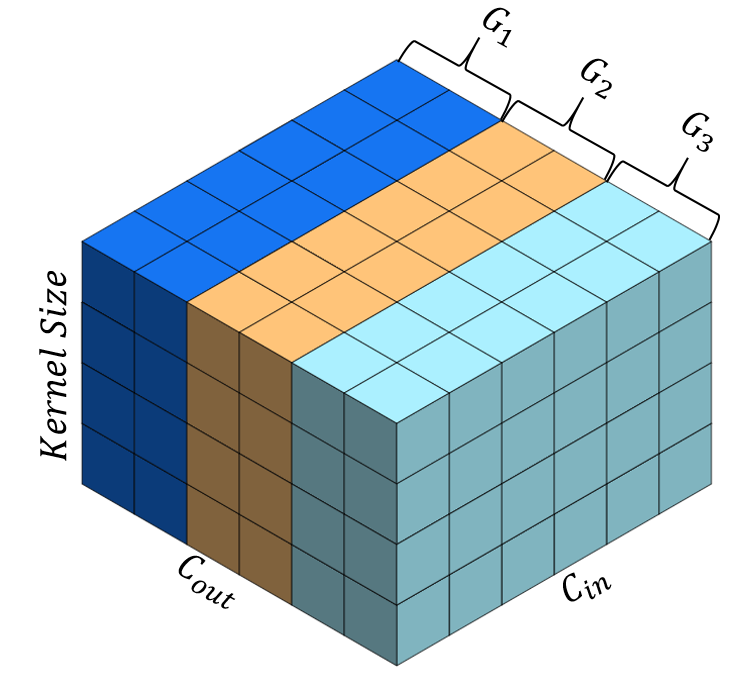}
\caption{Visualization of GOR's group partition for $N=3$. GOR enforces orthonormal regularization on groups of weights in the network layers. Best viewed in color.}
\label{fig:fig1}
\end{figure}

\section{Related Work}
{\bf Feature regularization} techniques aim to reduce the correlation between features in the same layer of a neural network. Batch normalization (BN) \cite{bn} and group normalization (GN) \cite{gn} are two widely used feature normalization techniques that have been shown to improve the performance of deep neural networks. BN normalizes the layer's activations to have zero mean and unit variance, while GN normalizes, for each example, the activations of a subset of channels within a given layer to have zero mean and unit variance. Alternatives to BN and GN include layer normalization \cite{layernorm} and instance normalization \cite{instancenorm}.

The DeCov approach \cite{DeCov} encourages diverse or non-redundant representations in deep neural networks by minimizing the cross-covariance of hidden activations. This method regularizes the feature maps of convolutional layers by encouraging them to be uncorrelated. In \cite{bank2018etf}, a relationship has been drawn between dropout and having an Extreme Value Theory Factorization (ETF) structure in auto-encoders. It has been suggested that imposing such a structure on neural network learning can improve learning.

{\bf Weight regularization} techniques aim to scale the weights of neural networks to have unit norm or zero mean and unit variance. Weight normalization \cite{salimans2016weight} is a technique that normalizes the weight vectors to have unit $\ell_2$ norm. 
Weight standardization (WS) \cite{qiao2019micro} is a method that attempts to smoothen the loss landscape by standardizing the weights (i.e., shifting them to have zero mean and unit variance) in convolutional layers. WS has been shown to improve the generalization performance of deep neural networks.

Unlike the reparametrization techniques listed above, which operate on each weight independently, the methods introduced in \cite{xie2017all, bansal2018can} perform regularization according to relations between filters within the layer. Specifically, they explore orthogonal regularization by promoting the Gram matrix of each weight matrix to be close to identity under the Frobenius norm.
The approach presented in \cite{li2022towards} proposes to impose orthogonality on specific network components, based on methods from sparse dictionary learning. 
Additional works suggested to enforce orthogonality on the filters \cite{ozay2016optimization,jia2017improving} which require computing their singular value decomposition.
In order to reduce the number of computations, it was further suggested to compute the singular value decomposition of groups of filters \cite{huang2018orthogonal}. Recently, a framework presented by \cite{su2022scaling} enforces strict orthogonality by using orthogonal filter bank parametrization and demonstrates the equivalence between different orthogonal convolutional layers in the spatial domain and the para-unitary systems in the spectral domain.

Our proposed method, Group Orthogonalization Regularization (GOR), is a weight regularization technique that promotes orthonormality between groups of filters within the same layer. GOR complements existing normalization techniques, such as BN and GN, and can be applied to a wide range of deep-learning layers and models. As enforcing orthogonality on all the weights in a given layer is quadratic in complexity, GOR enables increasing the layer dimensionality without sacrificing efficiency as it operates on groups and thus its complexity scales linearly with respect to the number of groups, and it scales quadratically only with respect to the group size that can be kept constant. Thus, it improves orthogonality between the weights while keeping the computational complexity reasonable. 




\section{Group Orthogonalization}
\subsection{Prelimenaries}
We start by describing some useful notation.
To train a DNN, the common practice is to use a loss function for learning a specific task. It is calculated according to the model parameters $W_{(1)}, ... , W_{(l)}$, and is denoted as $\mathcal{L}_{task}(W_{(1)}, ... , W_{(l)})$.
Note that $L$ is the number of layers in the model and $W_{(l)}$ the parameters of a specific layer ($l=1,...,L$).
The regularization we present in this work supports convolutional and fully connected layers.
Convolution layers are tensors of parameters, and we use their reshaped form. Let $\mathbf{W}_{(l)} \in \mathbb{R}^{w\times h\times c\times C_{out}}$ be the convolutional tensor, where $w$, $h$, $c$, $C_{out}$ are width, height, input channel number and output channel number, respectively.
We reshape $\mathbf{W}_{(l)}$ to be $W_{(l)} \in \mathbb{R}^{whc \times C_{out}}$. For simplicity, we define: $C_{in} = whc$.
Denote by $\norm{\cdot}_F$ the Frobenius norm of a matrix.

\subsection{Group Orthogonalization Regularization (GOR)} 
Similarly to \cite{bansal2018can}, we define \textit{orthogonalization} as the process of enforcing the Gram matrix of a set of filters to be close to identity according to a defined metric.

Previous works have proposed adding regularization on top of the task loss so that the Gram matrix of the parameters is approximately identity,
\begin{align}
\label{eq:reg_ortho}
    \mathcal{L}_{total}(W_{(1)}, ... , W_{(l)}) = \mathcal{L}_{task}(W_{(1)}, ... , W_{(l)}) + \lambda \sum_{l=1}^L \norm{W_{(l)}^T W_{(l)} - I}_F^2,
\end{align}
Yet, those methods may require an excessive amount of additional computation, especially for wide layers. A detailed explanation can be found in \cref{sec:computations}.

In this work, we suggest reducing the amount of computations by grouping the filters in each layer and orthogonalizing and normalizing only the filters within each group.
\begin{align}
\label{eq:GOR}
    \mathcal{L}_{total}(W_{(1)}, ... , W_{(l)}) = \mathcal{L}_{task}(W_{(1)}, ... , W_{(l)}) + \lambda \sum_{l=1}^L \sum_{i=1}^N \norm{W_{(i,l)}^T W_{(i,l)} - I}_F^2. 
\end{align}
Note that we have $N$ groups, each with $\frac{C_{out}}{N}$ filters in it. We flatten and stack the filters in each group to form a matrix $W_{(i,l)}$ of the $i$th group. Then, we enforce orthogonality within each of these groups.
The partition into groups is visually demonstrated in \cref{fig:fig1}.

In this way, we do not promote all the filters in a layer to be orthogonal to each other, which might be too restrictive, but rather orthogonal only to a subset of the filters. This can be beneficial, especially when $C_{in} << C_{out}$. In this case, $W_{(l)}^T W_{(l)} \in C_{out} \times C_{out}$ is necessarily not orthogonal and the rank of the Gram matrix is bounded by $C_{in}$.
In this case, using \cref{eq:reg_ortho} leads to a large magnitude of regularization with limited ability to orthogonalize the parameters.
With the partition into groups, this problem is smaller because we now penalize matrices with smaller output dimensions, which can be close to rank $C_{in}$ more easily.

\subsection{Computational Efficiency} \label{sec:computations}
A prominent advantage of GOR is its reduction in computation compared to whole-layer regularization.
The complexity of using orthogonalization between all weights as in \cref{eq:reg_ortho} is $C_{out}^2 C_{in}$ per layer. 
Yet, computing the regularization with groups as in \cref{eq:GOR} requires only $O(\frac{C_{out}^2C_{in}}{N})$.
This can be further improved by parallelizing the matrix multiplication of each group within the layer. The bounds are calculated according to the naive school book algorithm.


\subsection{Relation to Group Normalization}
\label{sec:gn_relation}
In the general case, where no other partition is applied to the intermediate features and/or filters, the groups can be chosen arbitrarily without affecting the learned function. This is because the order of the filters is not significant to the trained model.

One prominent method used for reducing the computational overhead of training DNNs is GN \cite{gn}, which replaces BN operations.
For GN, the intermediate output features of the network are partitioned into groups, and each group is centered and normalized.
In this case, it is no longer correct to state that the partition of the filters in each layer is arbitrary.
In this work, we distinguish between two possible options for partitioning the filters.
The first, which we call ``inter-group'' regularization where we use filters that match the normalization group and apply the regularization within this group.
The second is ``intra-group'' regularization, where we choose to apply our regularization to filters that match features from different groups of normalization. In particular, each filter in a regularization group corresponds to a different normalization group. 
The ``intra-group'' regularization may implicitly amplify the orthogonality between the normalization groups, while the ``inter-group'' regularization enhances orthogonality within normalization groups. Both settings are illustrated in \cref{fig:inter_intra} and further explained in \cref{sec:group_partition}.

\subsection{Group Partition}
\label{sec:group_partition}
Given the number of groups, $N$, we divide the kernel along the number of output channels, $C_{out}$, to $N$ consecutive groups - each contains $G = \frac{C_{out}}{N}$ filters. 
Given a partition, we distinguish between two cases:
\begin{enumerate}[label=(\alph*)]
\item Inter-Group regularization: The orthonormality is enforced on filters within the same group. Thus the regularization term (for a single layer) is
$\sum_{i=1}^N ||W_i^TW_i - I ||_F^2$, where 
\begin{align*}
    W_i^T = \left( f_{(i-1)G+1} | f_{(i-1)G + 2} | \ldots |f_{iG}  \right).
\end{align*}

\item Intra-Group regularization: The orthonormality is enforced on filters between different groups. In this case, the per-layer regularization term is 
$\sum_{i=1}^G ||W_i^TW_i - I ||_F^2$, where
\begin{align*}
    W_i^T = \left( f_i | f_{G + i} | f_{2G + i} |\ldots |f_{(N-1)G + i}  \right).
\end{align*}
\end{enumerate}

We focus on Inter-Group regularization.

\begin{figure}[h]
\centering
\includegraphics[scale=0.62]{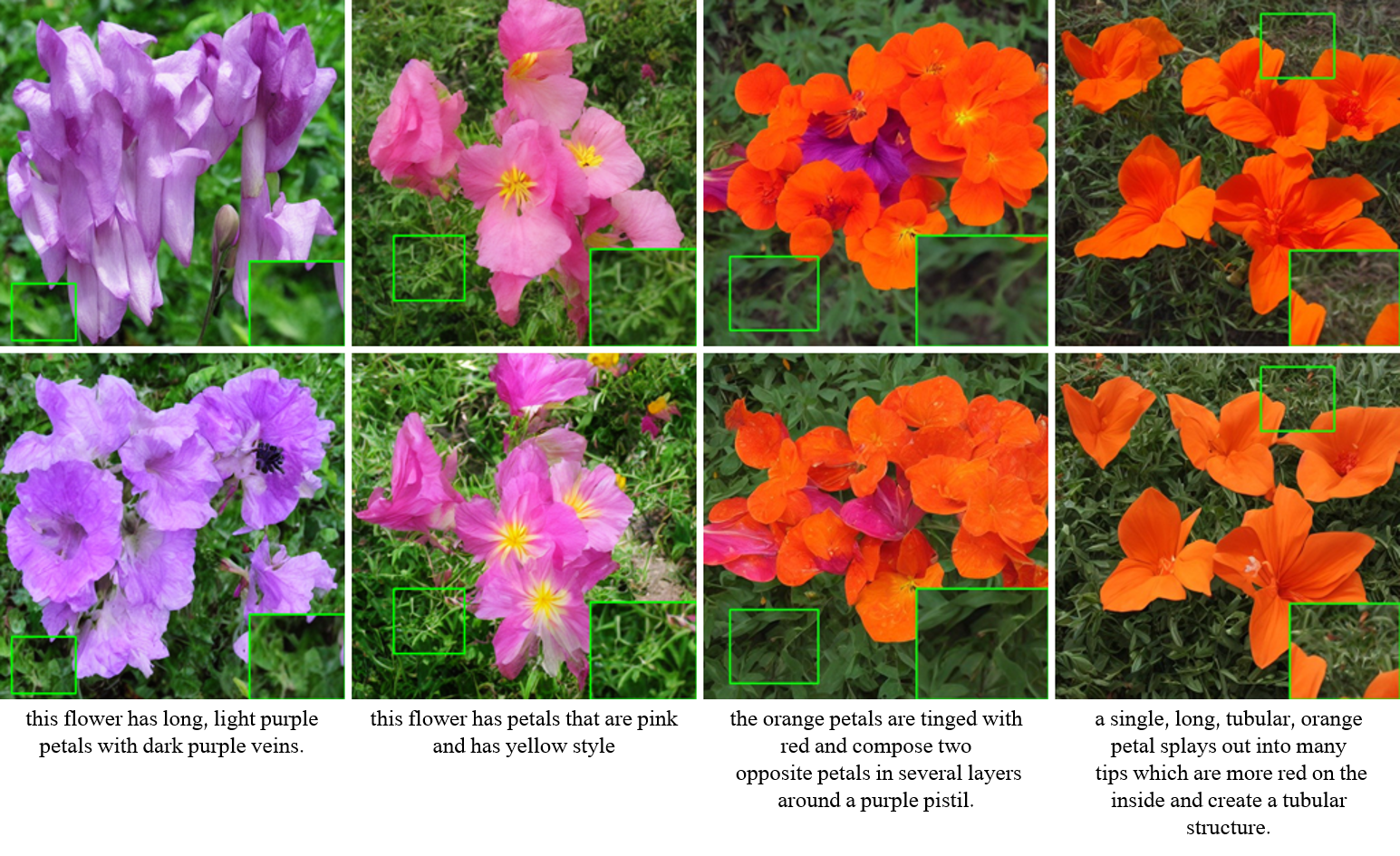}
\caption{Qualitative comparisons on Oxford102 between baseline fine-tuned model and model fine-tuned along with GOR using the same seed. The green rectangle is zoomed in by a factor of $1.5$. For each of the two rows: Top is LoRA baseline. Bottom is LoRA with our method. For the generation of the flowers themselves, the two models are comparable with similar artifacts, while our model is more successful at generating the background grass. This may be explained by the fact that we encourage orthogonality in the weights, which helps support more details.}
\label{fig:flowers_crop_grid}
\end{figure}


\subsection{GOR with Adapters}

A common paradigm in the field of deep learning is the use of foundation models  \cite{caron2021emerging,ramesh2021zero,Saharia2022Photorealistic,clip,llama,kirillov2023segment}. It relies on large-scale pretraining on general domain data which is followed by adaptation to particular tasks or domains. As models become larger, full fine-tuning becomes less feasible due to the heavy computation and the large storage needed. 
Some works that were introduced for large language models \cite{hu2021lora, mahabadi2021compacter, houlsby2019parameterefficient}, propose using lightweight adapters as an alternative to full model fine-tuning. One of the most prominent works for this task is LoRA \cite{hu2021lora}. 
Such adaptation approaches have also been proposed for the computation vision domain. The AdaptFormer \cite{dosovitskiy2020image} is a recent strategy that has shown promising results in the computer vision domain, especially when combined with vision transformers \cite{dosovitskiy2020image}.

The motivation for using GOR for model adaptation is that models trained with it are more likely to explore unknown directions in the parameter space, which may be beneficial when transforming between domains.

In our work, we focus on low-rank adapter modules as presented in \cite{hu2021lora, chen2022adaptformer}. Note that the AdaptMLP block follows the same decomposition as in LoRA but with a larger rank (64 vs. 4) and a non-linear layer in between. Both of these techniques fine-tune the foundation model weights by learning the residual that should be added to some selected layers. This residual is parameterized by a low-rank matrix that is composed of two matrices. The first reduces the size (`down sampling block'), and the second increases it (`up sampling block').  

GOR is applied to the adapter's matrices to encourage weights' expressibility.
Utilizing parameter space is more crucial for model performance in low-rank adapters.
Regularizing the downsampling block, i.e., the matrix that transforms from a high dimension to a small one, has little effect as there are few filters in each group.
Additionally, due to the extremely small output dimension of this block, it is most likely that features span the whole space. 
Thus, we only focus on applying GOR to the up-sampling block of the adapter.

\begin{table}
  \begin{center}
    \caption{CIFAR-10 Top-1 accuracy. ResNet110 GN is a 110-layer ResNet where all BN layers are replaced with GN. Both ResNet110 models are comprised of the basic block as the building block.}
    \label{tab:table1}
    \begin{tabular}{l c c c c}
    \toprule
      \textbf{Model} & \textbf{CE} & \textbf{SO} & \textbf{GOR inter} & \textbf{GOR intra} \\
      \midrule
      ResNet110                 & 93.95 $\pm$ 0.04 & \textbf{94.44 $\pm$ 0.05}    & 94.23 $\pm$ 0.07     & -\\
      ResNet110 GN                 & 92.02 $\pm$ 0.04 & 92.33 $\pm$ 0.03 & \textbf{92.73 $\pm$ 0.03}  & 92.24 $\pm$ 0.02 \\
    \bottomrule
    \end{tabular}
  \end{center}
\vspace{-0.15in}
\end{table}

\section{Experiments}
%
We evaluated the effectiveness of GOR regularization by conducting extensive experiments across several architectures and tasks. For each section, we describe the experimental setting and training details. 
\noindent  \subsection{Image Classification on CIFAR10}
To evaluate our method, we first perform image classification experiments.

\noindent  \textbf{Expermintal setting}. Using the 110-layer ResNet \cite{he2015deep} as a backbone, we benchmark our approach on the CIFAR-10 \cite{krizhevsky2009learning} dataset. To assess how well our technique complements GN, we also train the aforementioned architecture in its GN variant - meaning we replace all BN layers with GN layers. The rest of the model stays untouched. GOR is added to the optimization of all convolutional layers. Accuracy is compared to two baselines: non-regularized training (CE) and Soft-Orthogonalization (SO) \cite{bansal2018can}, which enforces the entire ($N = 1$ groups) set of filters to be orthonormal. For the GN variant, we tested both ``inter-group'' and ``intra-group'' GOR. Only ``inter-group'' is reported for the BN variant since, for this type of normalization, the two GOR variants are equivalent. As discussed in \cref{sec:gn_relation}.

\noindent  \textbf{Training details}. We train on the 50,000 training images set and evaluate on the 10,000 images test set. Top-1 accuracy is reported. With regard to batch size, number of epochs, weight decay, etc., training follows the exact same protocol as in \cite{he2015deep}, both for the GN model and the original model. Let $G$ be the number of groups for the GN layer. Following \cite{qiao2019micro}, the value of $G$ is determined to be the minimum value between 32 and (\# of channels)/4. As for the regularization's hyperparameters, the value of $\lambda$ is set to $10^{-2}$, and $N$ (number of groups for GOR) follows the same logic as $G$. For each table entry, we report the mean and std across three different random seeds.

\noindent  \textbf{Comparing regularization methods}. We compare the model's accuracy when trained with the two different regularization methods in \cref{tab:table1}. Both regularization methods improve generalization ability as both are superior to training without orthogonal regularization. For the original BN model, GOR showed competitive results to SO while being more efficient. The results show the benefit of using GOR along with GN, where the regularization enhances orthogonality within normalization groups.

\subsection{Adapting with GOR}
\subsubsection{Adapting ViT with GOR for image classification}
In this section, we report accuracy on several image datasets when fine-tuning a pre-trained ViT via the AdaptMLP \cite{chen2022adaptformer} bottleneck model.


\begin{figure}[t]
\centering
\includegraphics[scale=0.62]{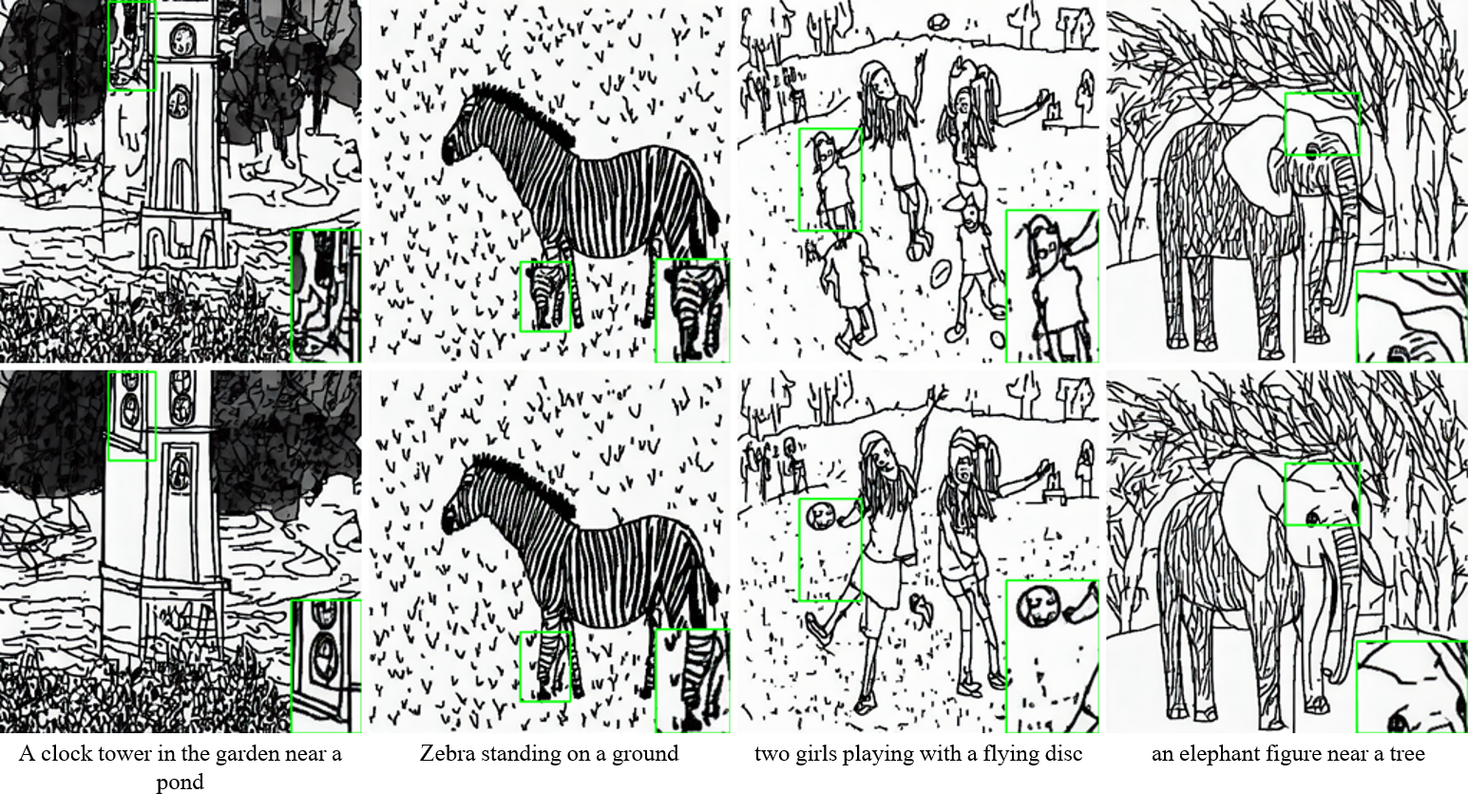}
\caption{Qualitative comparisons on FS-COCO between baseline fine-tuned model and model fine-tuned along with GOR using the same seed. The green rectangle is zoomed in by a factor of $1.5$. For each of the two rows: Top is LoRA baseline. Bottom is LoRA with our method. Our method improves the generation quality by both aligning with the text prompt more closely (second image from the right) and by removing artifacts.}
\label{fig:sketch_crop_grid}
\end{figure}

\noindent  \textbf{Expermintal setting}. Following experiments done by \cite{chen2022adaptformer}, we fine-tune two pre-trained Vision Transformers (both supervised and self-supervised) on multiple downstream datasets. The same ViT backbone model is used. We compare our results with a full fine-tuning setting and non-regularized AdaptFormer. In each training sequence, the pre-trained weights are frozen, and only the newly added modules are optimized. GOR regularization is applied to the AdaptMLP modules, specifically only to the up-projection layers. 

\noindent  \textbf{Datasets}. We use three common datasets: CIFAR-100 \cite{krizhevsky2009learning} contains 50,000 training images and 10,000 test images
of resolution $32\times 32$ with 100 labels. Street View House Numbers (SVHN) \cite{Netzer2011ReadingDI} is a digit classification
benchmark dataset. In total, the dataset comprises over 600,000 labeled images containing 73,257
training samples, 26,032 testing samples, and 531,131 extra training data. The Food-101 \cite{kaur2017combining} dataset
consists of 101 food categories with a total of 101k images, including 750 training and 250 testing samples per category.

\noindent  \textbf{Training details}. We follow the protocol described in the original work and use the same weights for the supervised and self-supervised baselines. The bottleneck dimension is set to 64 as recommended in the original paper \cite{chen2022adaptformer}. As for GOR configuration, We set $N$ to 16 and $\lambda$ to $10^{-4}$. Training is performed on 4 $\times$ NVIDIA GeForce RTX 2080 Ti GPUs together with gradient accumulation to match the original works' batch size.

\begin{table}
  \begin{center}
    \caption{Fine-tuning self-supervised ViT-B via AdaptMLP. Full-tuning and AdaptMLP results are taken from the original work. Pre-trained weights are from MAE \cite{he2022masked}. We report mean and std across 3 runs.}
    \label{tab:table2}
    \begin{tabular}{l c c c}
    \toprule
      \textbf{Method} & \textbf{CIFAR-100} & \textbf{SVHN} & \textbf{Food-101} \\
      \midrule
        Full-tuning                 & 85.9 & 97.67  & 90.09 \\
        AdaptFormer              & 85.9 & \textbf{96.89} & 87.61 \\
        AdaptFormer + GOR        & \textbf{86.16} $\pm$ 0.07 & 96.87 $\pm$ 0.09 & \textbf{87.73} $\pm$ 0.1\\
        \bottomrule
    \end{tabular}
  \end{center}
  \vspace{-0.15in}
\end{table}

\begin{table}
  \begin{center}
    \caption{Fine-tuning supervised ViT-B via AdaptMLP. Full-tuning and AdaptMLP results are taken from the original work. Pre-trained weights are from the ImageNet-21k \cite{deng2009imagenet} supervised pre-trained model. We report mean and std across 3 runs.}
    \label{tab:table3}
    \begin{tabular}{l c c c}
    \toprule
      \textbf{Method} & \textbf{CIFAR-100} & \textbf{SVHN} & \textbf{Food-101} \\
      \midrule
        Full-tuning              & 89.12 & 95.41  & 90.96 \\
        AdaptFormer              & 91.86 & 97.29 & 90.89 \\
        AdaptFormer + GOR         & \textbf{92.49} $\pm$ 0.11 & \textbf{97.36} $\pm$ 0.04 & \textbf{91.17} $\pm$ 0.02 \\
        \bottomrule
    \end{tabular}
  \end{center}
  \vspace{-0.1in}
\end{table}
        
\begin{table}[t]
  \begin{center}
    \caption{FID score comparisons on downstream datasets. Lower is better.}
    \label{tab:table4}
    \begin{tabular}{l c c c}
    \toprule
      \textbf{Method} & \textbf{Oxford102}   & \textbf{Pokemon-BLIP}  & \textbf{FS-COCO}\\
      \midrule
        LoRA               & 11.01 & 13.6 & 30.75\\
        LoRA + GOR         & \textbf{10.57} & \textbf{13.14} & \textbf{30.29} \\
        \bottomrule
    \end{tabular}
  \end{center}
  \vspace{-0.2in}
\end{table}

\noindent \textbf{Results}. \cref{tab:table2} and \cref{tab:table3} show that indeed our regularization improves the expressiveness of the bottleneck module and thus allows it to generalize better and achieve higher accuracy. GOR improves fine-tuning performance across almost all datasets and pre-trained weights. 

\subsubsection{Adapting Diffusion models with GOR for text-to-image generation}

We further evaluate the effectiveness of applying GOR to adapters by regularizing the fine-tuning process of image-generation models adapted with LoRA \cite{hu2021lora}. Efficient fine-tuning of diffusion models by injecting LoRA bottleneck modules into the U-Net's cross-attention layers was first proposed by \cite{Simo2022Repo}. While originally used to fine-tune large language models (LLMs) on downstream tasks, the writer suggested applying the rank-decomposition matrices to image generation models.  To our best knowledge, our work is the first time diffusion models adapted with LoRA are quantitively evaluated on the task of text-to-image generation.

\noindent  \textbf{Expermintal setting}. We measure how well the diffusion model adapts to the target data distribution when using GOR to regularize the LoRA matrices. The generated image fidelity of the adapted text-to-image model is evaluated using the Fréchet inception distance (FID) \cite{heusel2017gans} metric. For each dataset, the distance is computed between the generated images and all available real images. To reduce the impact of random variation, we compute FID three times in each experiment and report the minimum. LoRA \cite{hu2021lora} matrices are added to all four ($W_q$, $W_k$, $W_v$, $W_o$) of the U-Net's self-attention modules. As previously mentioned, during the model's optimization, we only enforce orthogonality via GOR on the "up" (denoted as $B$ in the original paper) matrices. Furthermore, we only regularize the adapter added to the upsampling blocks of the U-Net.

\noindent  \textbf{Datasets}. The Oxford 102 category flower dataset \cite{nilsback2008automated} consists of 102 flower categories. Each class consists of between 40 and 258 images. Images have large scale, pose, and light variations. The second dataset used is Pokemon-BLIP \cite{pinkney2022pokemon}. This image dataset consists of 833 text-image pairs. The pairs include a Pokemon image and a BLIP-generated \cite{li2022blip} caption. FS-COCO comprises 10,000 freehand scene vector sketches drawn by 100 non-expert individuals, offering both object and scene-level abstraction. Each sketch is augmented with its text description. Representative examples can be seen in \cref{fig:fs_coco_vis}.

\noindent  \textbf{Note on FID calculation}. When measuring FID on the Oxford102 and FS-COCO datasets, we use 10,000 generated images to ensure that representative statistics are calculated. Due to the small size of the Pokemon-BLIP dataset, we only use 583 generated images. For all datasets, we randomly sample prompts and use them to condition the diffusion model during generation. More information on measuring the FID is listed in \cref{sec:diff_detail}. 

\noindent  \textbf{Training details}. Pre-trained latent diffusion model weights taken from the HuggingFace Diffusers library \cite{von-platen-etal-2022-diffusers}. Specifically, the \textit{stable-Diffusion-v1-5} checkpoint is used.  Following the original work, we set the LoRA rank ($r$ from paper) to 4 and only train the adapter's matrices while keeping the rest of the weights frozen. We set $N = 32$ to match the groups of normalized features in the GN layer and train the model for 15K iterations. The value of $\lambda$ is set to $10^{-5}$ for the Oxford102 dataset and $10^{-6}$ for the other two datasets. All experiments are conducted on 2 $\times$ NVIDIA GeForce RTX 2080 Ti. The rest of the training protocol is listed in the \cref{sec:diff_detail}.

\noindent \textbf{Effect of regularization on FID score}. \cref{tab:table4} shows that enforcing group-orthogonality of the LoRA layers throughout the model's fine-tuning improves its ability to learn the underlying distribution. \cref{fig:flowers_crop_grid,fig:pokemon_crop_grid,fig:sketch_crop_grid} show a qualitative comparison between GOR and the baseline.

\subsection{Adversarial Robustness}
\label{sec:adv_robust}
A major challenge faced by deep neural networks is their vulnerability to small changes in input data, which can result in incorrect predictions. This presents a significant difficulty, particularly for applications that require high safety standards. To tackle this issue, the development of adversarial defenses has emerged as a critical research area across various fields, including machine learning, computer vision, natural language processing, and others.

One of the approaches utilized to defend against such attacks is adversarial training \cite{madry2017towards, zhang2020attacks, jia2022adversarial, zhang2019theoretically, shafahi2019adversarial}. It involves generating adversarial examples during the training process and utilizing them to update the model parameters. The adversarial training scheme can be viewed as a form of regularization that prevents the model from overfitting to the distributions of clean training data.

Recent works \cite{xu2022orthogonal, jalwana2020orthogonal} have shown that model orthogonality improves robustness. Motivated by these results, we investigate how adversarial training methods can benefit from our efficient group orthogonalization.

\textbf{Expermintal setting}. We evaluate the robustness of two adversarial training methods when combined with GOR on CIFAR-10 in \cref{tab:adv_table}. The Trained model is evaluated under several white-box and black-box attacks. The two well-known training techniques we consider are TRADES \cite{zhang2019theoretically} and FAT \cite{zhang2020attacks}. We train both GN and BN models.

\textbf{Adversarial Attacks} The adversarial test samples are bounded by $L_\infty$ perturbations with $\epsilon = 8 / 255$, which are generated by FGSM, PGD$_{20}$, PGD$_{100}$ and CW$_\infty$.
Where the subscript numbers indicate the number of iterations used for calculating the attack and
CW$_\infty$ is the $L_\infty$ version of C\&W loss \cite{carlini2017towards} optimized by PGD$_{20}$. 
Trained models are also attacked by the AutoAttack \cite{croce2020reliable} method, which consists of APGD-CE, APGD-DLR, FAB \cite{croce2020minimally} and Square \cite{andriushchenko2020square}. 

\textbf{Training and evaluation details}. We follow the original settings for each of the training methods, with Wide ResNet \cite{zagoruyko2016wide} as the chosen architecture. 
We train a WideResNet-34-10 for TRADES and a WideResNet-32-10 for FAT. The GN model is created by replacing all BN layers with GN, with $G = 32$. We report the test accuracy of the deep model at the last training epoch (76 for TRADES and 120 for FAT). As for GOR configuration, we keep $N = G = 32$ and show results for $\lambda$ at $10^{-4}$ and $10^{-5}$ for all models. Inter-group regularization is used.

\textbf{Effect of regularization on robustness}. \cref{tab:adv_table} justifies the assumption that GOR allows the model to learn more diverse and robust features that are less correlated and more informative. Models regulated by our methods show better natural accuracy and, in most cases, are more robust to attacks. The gains are more significant for the GN models. This affirms our claim that GOR complements GN well, especially when regularizing according to the normalized groups.

\begin{table}
  \begin{center} 
    \caption{Test accuracy under different training methods on CIFAR10 dataset under attacks bounded by $L_\infty$. 34-layer wide ResNet is used for TRADES \cite{zhang2019theoretically} training, and a 32-layer wide ResNet is used for FAT training \cite{zhang2020attacks}.}
    \label{tab:adv_table}
    \adjustbox{width=0.8\textwidth}{
    \begin{tabular}{l c c c c c c c c}
    \toprule
      \textbf{Method} & \textbf{$\lambda$} & \textbf{Natural} 
        & \textbf{FGSM} & \textbf{PGD$_{20}$}
        & \textbf{PGD$_{100}$} & \textbf{CW$_{\infty}$} & \textbf{AutoAttack}\\
      \midrule
        TRADES + BN & 0  & 83.73 & 66.12 & \textbf{55.85} & 55.52 & 53.55 & 52.45   \\
        TRADES + BN & $10^{-4}$ & \textbf{84.63} & \textbf{66.65} & 55.16 & 55.00 & 53.66 & 52.33  \\
        TRADES + BN & $10^{-5}$ & 84.01 & 66.5 & 55.13 & \textbf{55.93} & \textbf{54.33} & \textbf{53.14}  \\
        \rowcolor{Gray}
        TRADES + GN & 0 & 81.88 & 63.72 & 53.27 & 53.11 & 50.87 & 49.72\\
        \rowcolor{Gray}
        TRADES + GN & $10^{-4}$ & \textbf{83.86} & \textbf{65.67} & \textbf{54.76} & \textbf{54.58} & \textbf{52.87}  & \textbf{51.56}\\
        \rowcolor{Gray}  
        TRADES + GN & $10^{-5}$ & 82.72 & 64.42 & 54.04 & 53.77 & 51.77 & 50.66 \\
        \midrule
        FAT + BN & 0 & 88.6 & \textbf{65.05} & 46.34 & 45.5 & 46.87 & 43.62 \\
        FAT + BN & $10^{-4}$ & 88.83 & 64.49 & 45.82 & 45.1 & 46.58 & 43.39 \\
        FAT + BN & $10^{-5}$ & \textbf{88.85} & 65 & \textbf{46.72} & \textbf{45.83} & \textbf{47.16} & \textbf{43.71} \\
        \rowcolor{Gray}
        FAT + GN & 0 & 82.83 & 54.13 & 36.34 & 35.68 & 38.47 & 34.19  \\
        \rowcolor{Gray}
        FAT + GN & $10^{-4}$ & 87.87 & 64.84 & 46.31
        &  45.77 & 47.34 & 44.52 \\
        \rowcolor{Gray}
        FAT + GN & $10^{-5}$ &\textbf{ 88.17} & \textbf{66.73} & \textbf{47.07} & \textbf{46.4} & \textbf{48.37} & \textbf{45.04} \\
    \bottomrule
    \end{tabular}  
}
  \end{center}
\end{table}

\subsection{Computational Efficency}
\begin{figure}[t]
  \centering
  \includegraphics[width=0.8\linewidth]{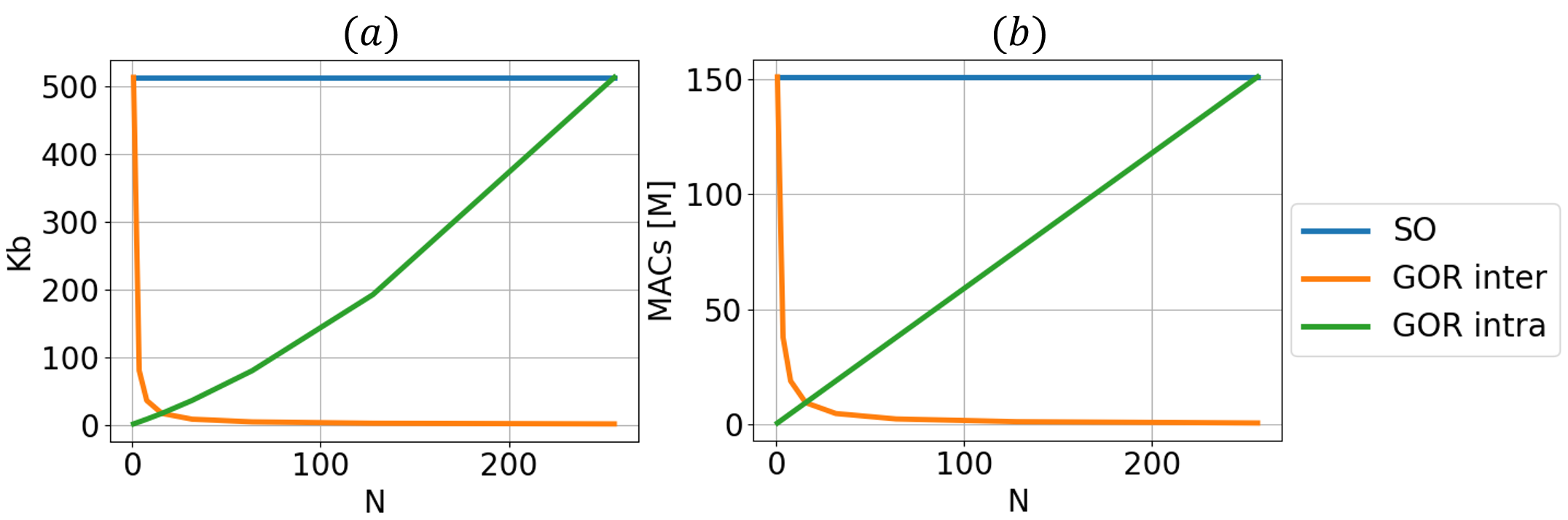}
   \caption{For different $N$ (group size) values, we report (a) runtime, (b) multiply-accumulate (MAC). GOR improves over SO in terms of MACs and memory while getting accuracy improvement.}
   \label{fig:reg_comparison}
\end{figure}

\cref{fig:reg_comparison} include a comparison of the regularization methods across memory consumption and number of operations. In each experiment, we calculate the regularization term for a single convolution layer with a kernel of dimensions: $C_{out} \times C_{in} \times h \times w = 256 \times 256 \times 3 \times 3$. We compare non-grouped orthogonalization regularization (SO) and the two GOR variants. All experiments were performed on NVIDIA GeForce RTX 2080 Ti.

\subsection{Limitations} 
While GOR shows promising results on multiple tasks, some limitations need to be addressed in future work. First, it introduces a new hyper-parameter, $\lambda$, that might need to be re-tuned depending on the architecture and task. For non-GN models, $N$ might need to be tuned as well. Second, and similarly to other orthogonalization regularizations \cite{bansal2018can, xie2017all}, it introduces some computational overhead that is proportional to the number of regularized layers.

%

\section{Conclusion}

In this study, we propose a novel regularization technique that encourages orthonormality between groups of filters within the same layer. This technique is computationally efficient and can significantly reduce the redundancy within the parameters of deep neural networks.

Our experiments show that incorporating our regularization technique into recent adaptation methods for diffusion models and vision transformers (ViTs) leads to improved performance on downstream tasks. Moreover, enforcing group orthogonality during adversarial training results in better model robustness.
Overall, our proposed regularization technique effectively enhances the performance and robustness of deep neural networks. By reducing the redundancy within the network's parameters, we can create more efficient and accurate models that perform better on a variety of tasks.

\paragraph{Acknowledgements}
This work was partially supported by the European research council under Grant ERC-StG 757497.

\bibliography{ref}


\renewcommand\thesection{\Alph{section}}
\renewcommand\thesubsection{\thesection.\arabic{subsection}}
\setcounter{section}{0}

\section*{Appendix}

\section{Ablation Study}
In this section, we conduct an ablation study to assess the impact of different hyperparameters on the effectiveness of our proposed weight regularization technique. Specifically, we examine the effects of the orthogonalization group size, its interaction with GN, and the magnitude of regularization.
For all the experiments in this section, we maintain a consistent configuration, except for the parameter under investigation, which is varied accordingly. We utilize the ResNet110 architecture with GN and incorporate inter-group GOR.

We report the top-1 accuracy for different values of $N$ (number of regularization groups), $G$ (number of normalization groups in the GN layer) and $\lambda$ (regularization strength) in \cref{tab:n_table,tab:g_table,tab:lambda_table} respectively. 

As mentioned in the main paper, we keep the number of filters/channels in each group to be at least 4, meaning that for every layer, the following holds:
$$ 
N_{(l)} = \min\{N, \ \frac{C_{out}}{4} \} \quad \text{and} \quad G_{(l)} = \min\{G, \ \frac{C}{4} \}.
$$
Due to this limitation, the neural networks utilized in this study consist of convolutional layers with a number of channels that allows the values of $N$ and $G$ to reach a maximal value of $16$.

\cref{tab:n_table} shows that optimal outcomes are achieved by aligning orthogonalization groups with the normalization group, i.e. $N=G$. 
This way, the orthogonality among the normalization groups increases.
\cref{tab:g_table} supports our choice of group size.
The results in \cref{tab:lambda_table} present the hyperparameter search for the optimal value of $\lambda$.

\begin{table}[h]
  \begin{center}
    \caption{CIFAR10 Top-1 accuracy for a varying number of groups, $N$. ResNet110 GN model is used. We keep the number of normalization groups to be $G$ = $\min$\{32, \#channels / 4\}. Mean and std across 3 seeds are reported.}
    \label{tab:n_table}
    \begin{tabular}{l c c c c c}
    \toprule
        \textbf{$N$} & \textbf{1} & \textbf{2} & \textbf{4} & \textbf{8} & \textbf{16}\\        \textbf{$G$} & \textbf{16} & \textbf{16} & \textbf{16} & \textbf{16} & \textbf{16}\\
      \midrule
         & 92.33 $\pm$ 0.03 & 92.55 $\pm$ 0.11 & 92.16 $\pm$ 0.03 & 92.31 $\pm$ 0.08 & \textbf{92.73 $\pm$ 0.03}\\
    \bottomrule
    \end{tabular}
  \end{center}
\end{table}

\begin{table}[h]
  \begin{center}
    \caption{CIFAR10 Top-1 accuracy for different values of $G$. ResNet110 GN model is used. We keep $N$ = $G$. Mean and std across 3 seeds are reported.}
    \label{tab:g_table}
    \begin{tabular}{l c c c c c c}
    \toprule
      \textbf{$N$} & \textbf{1} & \textbf{2} & \textbf{4} & \textbf{8} & \textbf{16}\\ 
      \textbf{$G$} & \textbf{1} & \textbf{2} & \textbf{4} & \textbf{8} & \textbf{16}\\
      \midrule 
        & 92.19 $\pm$ 0.17  & 92.45 $\pm$ 0.1 & 92.59 $\pm$ 0.17 & 92.45 $\pm$ 0.15 & \textbf{92.73 $\pm$ 0.03}\\
    \bottomrule
    \end{tabular}
  \end{center}
\end{table}

\begin{table}[h]
  \begin{center}
    \caption{CIFAR10 Top-1 accuracy for different values of $\lambda$. ResNet110 GN model is used. We report mean and std across 3 seeds.}
    \label{tab:lambda_table}
    \begin{tabular}{lc c c c c c}
    \toprule
      \textbf{$\lambda$}&\textbf{$10^{-1}$} & \textbf{$10^{-2}$} & \textbf{$10^{-3}$} & \textbf{$10^{-4}$} & \textbf{$10^{-5}$}\\
      \midrule 
        & 92.44 $\pm$ 0.17 & \textbf{92.73 $\pm$ 0.03} & 92.22 $\pm$ 0.21 & 92.53 $\pm$ 0.1 & 92.42 $\pm$ 0.4\\
    \bottomrule
    \end{tabular}
  \end{center}
\end{table}

\section{Inter vs. Intra GOR } 
\begin{figure}[h]
\centering
\includegraphics[scale=0.25,width=0.8\linewidth]{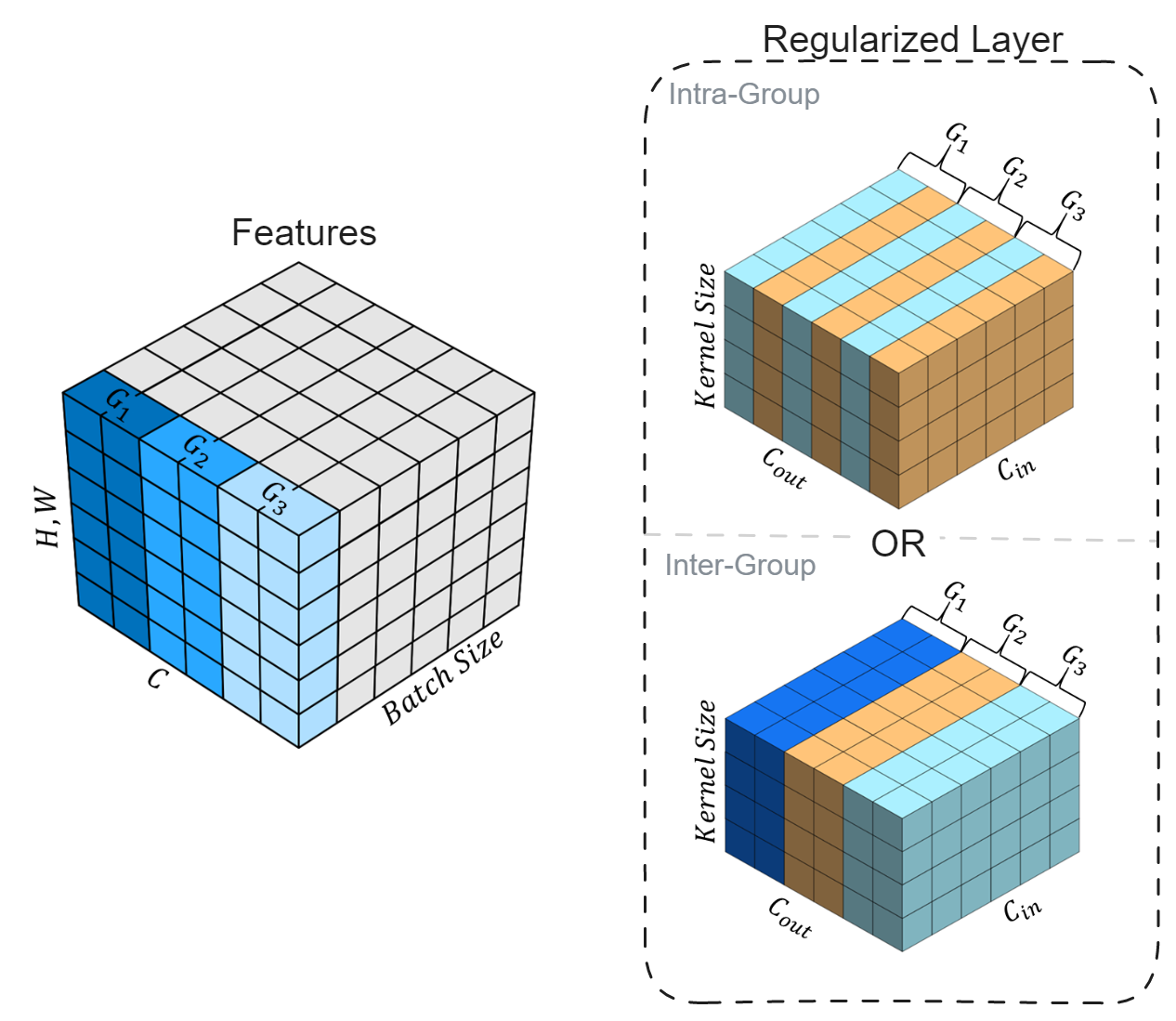}
\caption{Partition of filters for GOR according to Inter-Group and Intra-Group for $N = 3$. Input features (left) are colored according to GN normalization with $G = 3$. Filters (right) are colored according to the sets orthogonality is enforced on. Best viewed in color.}
\label{fig:inter_intra}
\end{figure} 

\cref{fig:inter_intra} visualizes the difference between ``inter'' and ``intra'' group partition with GN. 
The groups of filters are determined according to the normalization groups of the features.
As discussed in the paper, in the inter-group setting, filters within the same group are enforced to form an orthonormal set. On the other hand, in the intra-group setting, we enforce orthonormality between filters from different groups.

\section{Diffusion Models Adapters - Experiments Details}
\label{sec:diff_detail}
In this section, we elaborate on the training and evaluation protocols of adapters of diffusion models presented in Section 4.2.2 of the paper.

\textbf{Experiment setting}. Our training protocol is built upon the 
example\footnote{https://github.com/huggingface/diffusers/tree/main/examples/text\_to\_image} published by HuggingFace \cite{von-platen-etal-2022-diffusers}. For the Pokemon-BLIP dataset \cite{pinkney2022pokemon}, we train with a batch size of 4 and 512$\times$512 resolution. As for the Oxford102  \cite{nilsback2008automated} and the FS-COCO \cite{chowdhury2022fs} datasets, we use a batch size of 64 and 256$\times$256 resolution. We set the base learning rate to $10^{-4}$ and apply a cosine scheduler. Data is pre-processed using central crop and normalization. Random flip is employed as data augmentation.

\textbf{FID calculation}. The procedure consists of two stages: first, producing samples from the model; second, computing the discrepancy between the InceptionV3 \cite{szegedy2016rethinking} statistics of the model-generated images and the original ones. For both steps, we build upon the code published for \cite{karras2022elucidating}.
Following common practice, before being passed to the Inception model for statistics calculation, the images (both generated and non-generated) are undergone the same pre-processing (normalization and central crop) as mentioned above.


\begin{figure}[h]
\centering
\includegraphics[scale=0.7]{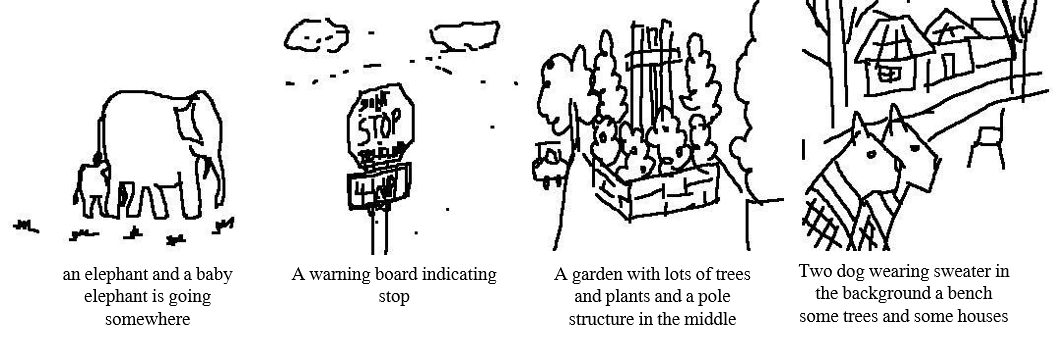}
\caption{Examples of image-text pairs from the FS-COCO dataset}
\label{fig:fs_coco_vis}
\end{figure}




\section{Qualitative Examples} 
\begin{figure}[h]
\centering
\includegraphics[scale=0.7]{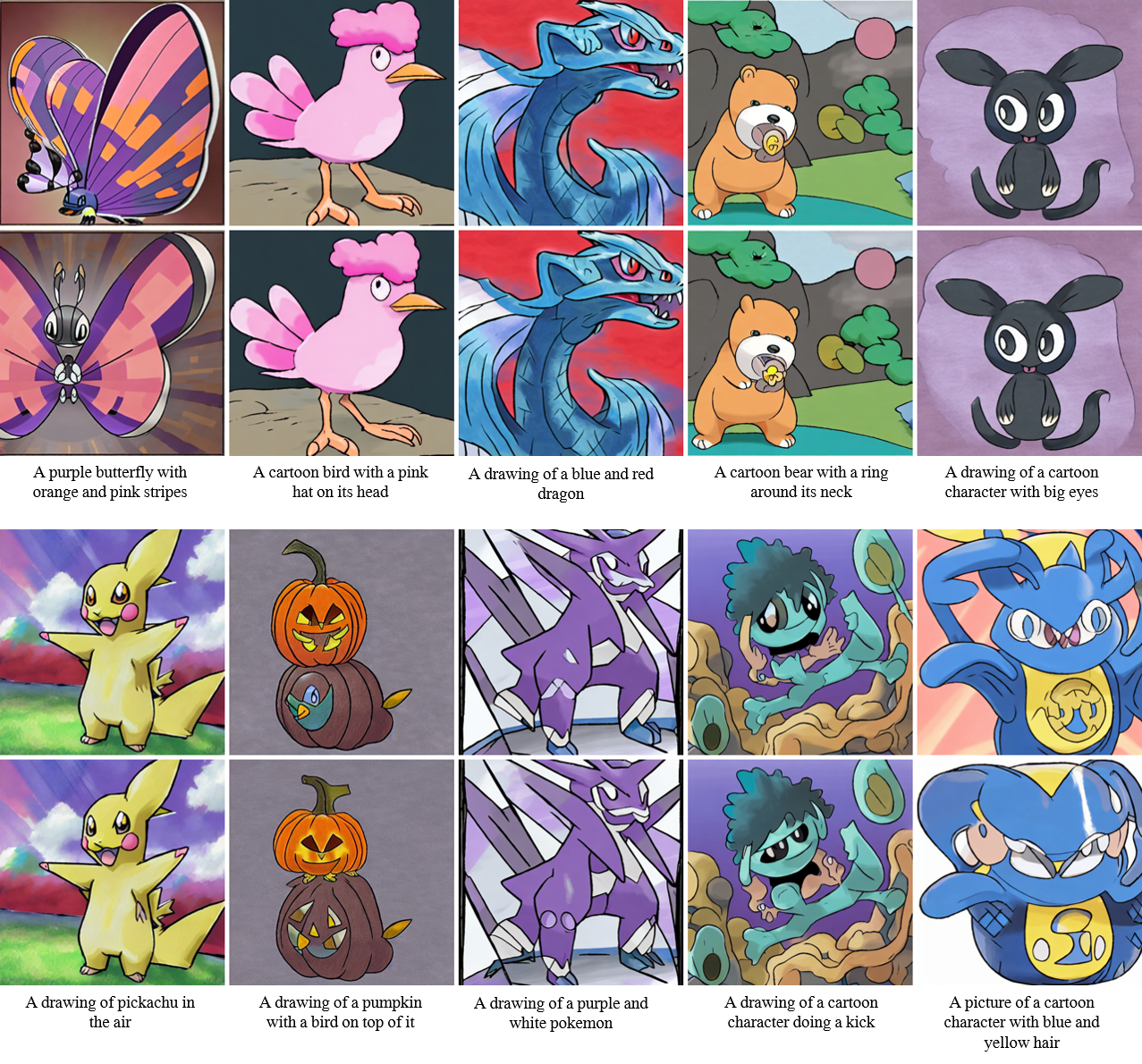}
\caption{Qualitative comparisons on Pokemon-BLIP between baseline fine-tuned model and model fine-tuned along with GOR using the same seed. For each of the two rows:  Top is LoRA baseline. Bottom is LoRA with our method.}
\vspace{-2cm}
\label{fig:pokemons_1}
\end{figure}

\begin{figure}[h]
\centering
\includegraphics[scale=0.7]{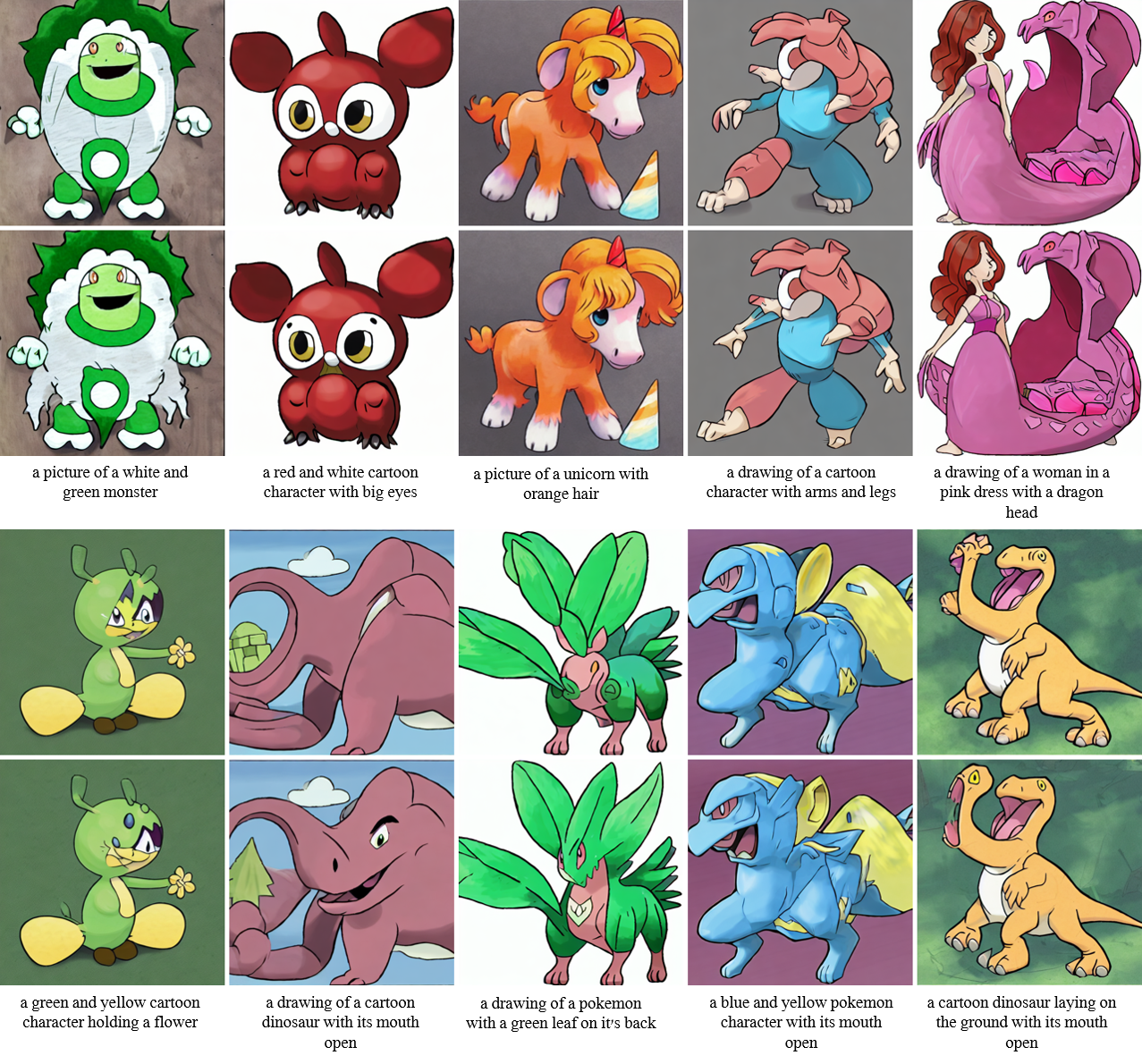}
\caption{Qualitative comparisons on Pokemon-BLIP between baseline fine-tuned model and model fine-tuned along with GOR using same seed. For each of the two rows:  Top is LoRA baseline. Bottom is LoRA with our method.}
\label{fig:pokemons_2}
\end{figure}

\begin{figure}[t]
\centering
\includegraphics[scale=0.7]{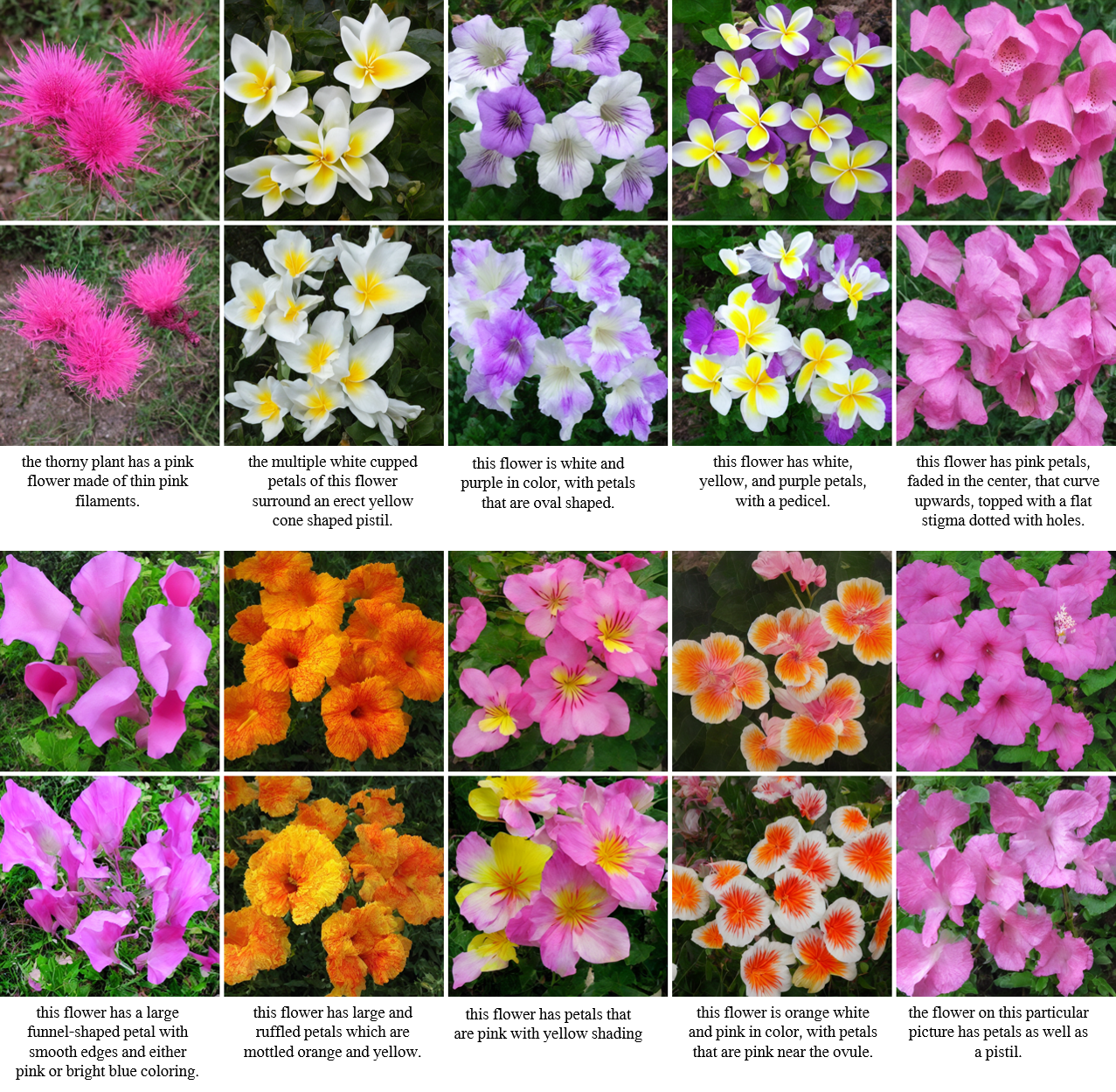}
\caption{Qualitative comparisons on Oxford102 between baseline fine-tuned model and model fine-tuned along with GOR using the same seed. For each of the two rows:  Top is LoRA baseline. Bottom is LoRA with our method.}
\label{fig:flowers_1}
\end{figure}

\begin{figure}[t]
\centering
\includegraphics[scale=0.7]{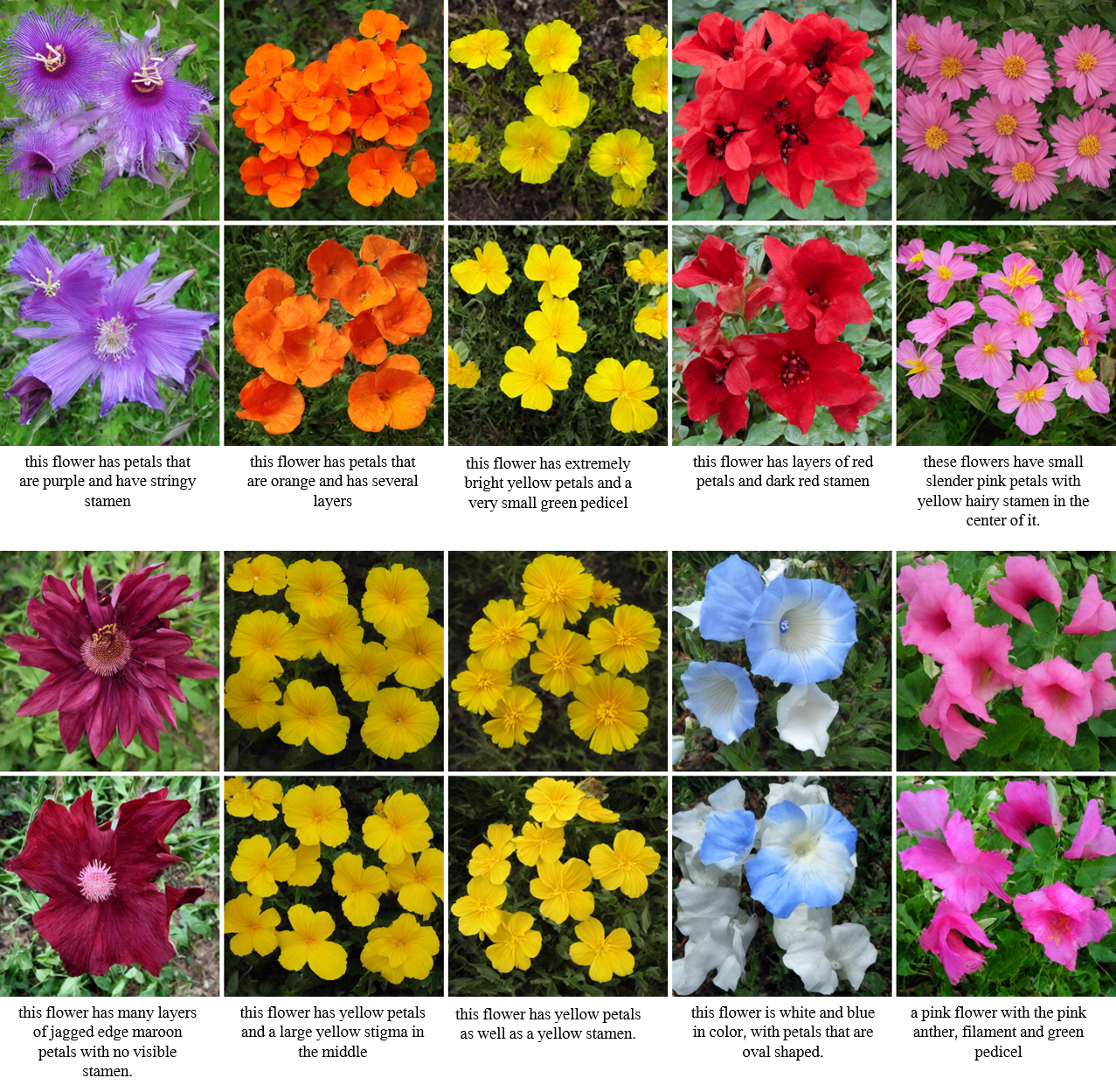}
\caption{Qualitative comparisons on Oxford102 between baseline fine-tuned model and model fine-tuned along with GOR using the same seed. For each of the two rows:  Top is LoRA baseline. Bottom is LoRA with our method.}
\label{fig:flowers_2}
\end{figure}

\begin{figure}[t]
\centering
\includegraphics[scale=0.7]{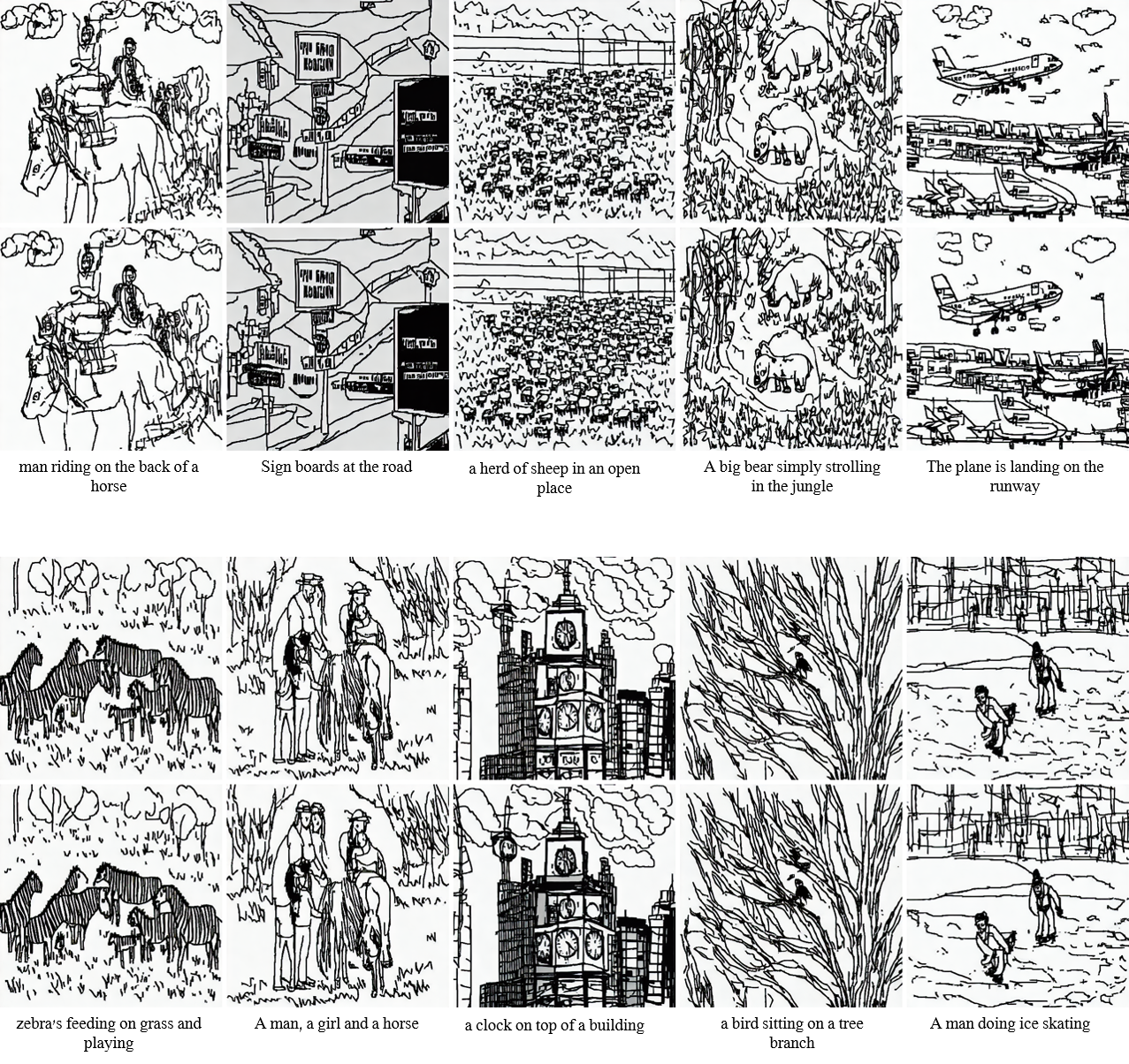}
\caption{Qualitative comparisons on FS-COCO between baseline fine-tuned model and model fine-tuned along with GOR using the same seed. For each of the two rows:  Top is LoRA baseline. Bottom is LoRA with our method.}
\label{fig:sketch_1}
\end{figure}

\begin{figure}[t]
\centering
\includegraphics[scale=0.7]{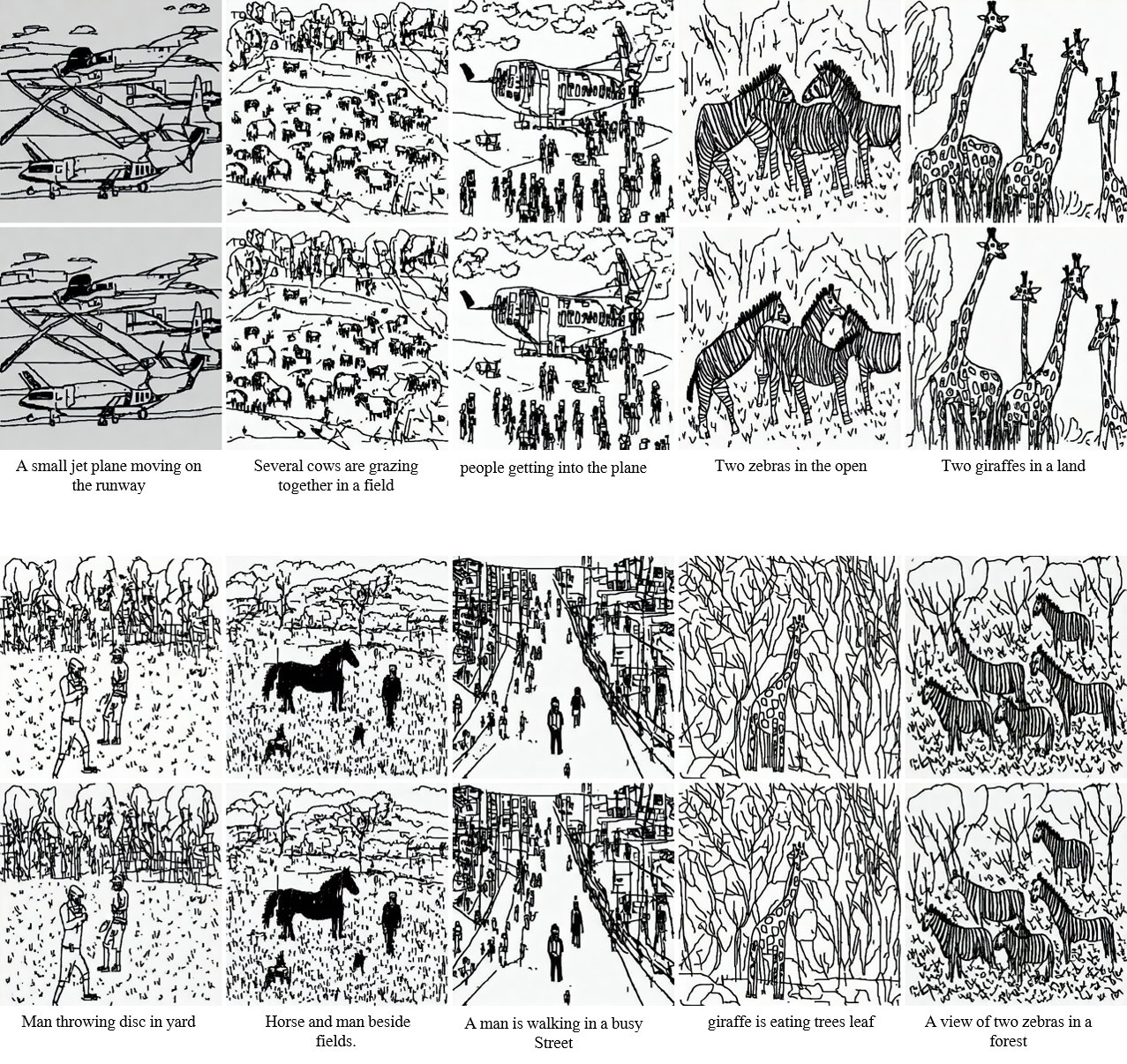}
\caption{Qualitative comparisons FS-COCO between baseline fine-tuned model and model fine-tuned along with GOR using the same seed. For each of the two rows:  Top is LoRA baseline. Bottom is LoRA with our method.}
\label{fig:sketch_2}
\end{figure}

We present more qualitative comparisons between our method and the baseline in \cref{fig:pokemons_1,fig:pokemons_2,fig:flowers_1,fig:flowers_2,fig:sketch_1,fig:sketch_2}. The text prompt used to condition the generative model is presented at the bottom of each pair. Note that the presented results are randomly generated with no cherry-picking.

\end{document}